\documentclass{article}

\usepackage{microtype}
\usepackage{graphicx}
\usepackage{subfigure}
\usepackage{booktabs} 

\usepackage{hyperref}

\usepackage[accepted]{icml2024}

\usepackage{amsmath}
\usepackage{amssymb}
\usepackage{mathtools}
\usepackage{amsthm}
\usepackage{bm}
\usepackage{multirow} 

\usepackage[capitalize,noabbrev]{cleveref}

\theoremstyle{plain}

\theoremstyle{definition}

\theoremstyle{remark}

\usepackage[textsize=tiny]{todonotes}

\icmltitlerunning{Counterfactual Reasoning for Multi-Label Image Classification via Patching-Based Training}

\begin{document}

\def \x {\bm{x}}
\def \z {\bm{z}}
\def \y {\bm{y}}
\def \o {\bm{o}}
\def \u {\bm{u}}
\def \p {\bm{p}}
\def \q {\bm{q}}
\def \w {\bm{w}}
\def \i {\bm{i}}

\twocolumn[
\icmltitle{Counterfactual Reasoning for Multi-Label Image Classification \\ via Patching-Based Training}



\icmlsetsymbol{equal}{*}

\begin{icmlauthorlist}
\icmlauthor{Ming-Kun Xie}{equal,nuaa}
\icmlauthor{Jia-Hao Xiao}{equal,nuaa}
\icmlauthor{Pei Peng}{nuaa}
\icmlauthor{Gang Niu}{riken}
\icmlauthor{Masashi Sugiyama}{riken,tu}
\icmlauthor{Sheng-Jun Huang}{nuaa}

\end{icmlauthorlist}

\icmlaffiliation{nuaa}{Nanjing University of Aeronautics and Astronautics, Nanjing, China}
\icmlaffiliation{riken}{RIKEN Center for Advanced Intelligence Project, Tokyo, Japan}
\icmlaffiliation{tu}{The University of Tokyo, Tokyo, Japan}

\icmlcorrespondingauthor{Sheng-Jun Huang}{huangsj@nuaa.edu.cn}

\icmlkeywords{Machine Learning, ICML}

\vskip 0.3in
]



\printAffiliationsAndNotice{\icmlEqualContribution} 

\begin{abstract}
	
The key to multi-label image classification (MLC) is to improve model performance by leveraging label correlations. Unfortunately, it has been shown that overemphasizing co-occurrence relationships can cause the overfitting issue of the model, ultimately leading to performance degradation. In this paper, we provide a causal inference framework to show that the correlative features caused by the target object and its co-occurring objects can be regarded as a mediator, which has both positive and negative impacts on model predictions. On the positive side, the mediator enhances the recognition performance of the model by capturing co-occurrence relationships; on the negative side, it has the harmful causal effect that causes the model to make an incorrect prediction for the target object, even when only co-occurring objects are present in an image. To address this problem, we propose a counterfactual reasoning method to measure the total direct effect, achieved by enhancing the direct effect caused only by the target object. Due to the unknown location of the target object, we propose patching-based training and inference to accomplish this goal, which divides an image into multiple patches and identifies the pivot patch that contains the target object. Experimental results on multiple benchmark datasets with diverse configurations validate that the proposed method can achieve state-of-the-art performance. The implementation is available at \url{https://github.com/xiemk/MLC-PAT}.

\end{abstract}

\section{Introduction}

In \textit{single-label} supervised learning, each instance is assigned with a single label to describe its semantics, while many realistic scenarios exhibit a \textit{multi-labeled} nature, where each instance is associated with multiple labels \cite{zhang2013review,liu2021emerging}. For example, an image captured on the beach may simultaneously contain objects such as \textit{sea}, \textit{person}, and \textit{palm tree}. The goal of MLC is to develop a classification model based on examples with multiple semantics, enabling it to predict all relevant labels for unseen instances. 


In comparison to single-label learning, tackling the MLC tasks proves significantly more challenging, mainly due to the exponentially expanded output space. Given the prevalent co-occurrence of objects in the real world, many methods have emerged to address the MLC problems by exploiting label correlations. In the context of deep learning, the pioneering work \cite{wang2016cnn} employed recurrent neural networks (RNNs) to capture high-order label correlations. Subsequent methods focused on integrating co-occurrence relationships into training, achieved through the design of specialized architectures, \textit{e.g.}, graph convolutional networks (GCNs) \cite{chen2019multi}, or the development of innovative training schemes,  \textit{e.g.}, label mask training \cite{lanchantin2021general}.



While these methods have improved the practical performance of MLC, an excessive emphasis on co-occurrence has been shown to misguide model training, leading to a degradation in overall performance \cite{xu2022boosting}. To address this problem, the recent work \cite{xu2022boosting} decomposed the original task into co-occurrence and dis-occurrence subtasks. This enabled the former to capture joint patterns, while the latter to focus on class-specific features. Another study \cite{liu2023causality} proposed to treat co-occurring objects as confounders, recognizing the potential for incorrect causalities between objects and predictions. To mitigate this issue, it performed an attention-based intervention to eliminate the harmful confounding effect.

In this paper, we conduct a systematic study to investigate the impact of label co-occurrence in MLC. We first reveal that when using the common binary cross entropy~(BCE) or its variant, the asymmetric loss~(ASL) \cite{ridnik2021asymmetric} loss, which does not explicitly consider co-occurrence relationships, the model inherently captures correlative patterns among objects co-occurring within a single image. Moreover, we provide a theoretical explanation of this phenomenon from a causal perspective. We find that the correlative features derived from the target object and its co-occurring objects can be regarded as a mediator, exerting both positive and negative influences on model predictions. On the positive side, it enhances the ability of the model to identify co-occurring objects by exploiting correlative patterns; on the negative side, it causes the model to overfit to correlative features, resulting in inaccurate predictions of non-existent objects due to the misguidance by their co-occurring ones.

To address this problem, our main idea is to mitigate the negative impact of the mediated effect, \textit{i.e.}, the indirect effect when the target object is masked yet the co-occurrence information remains activated due to the presence of co-occurring objects. This motivates us to access the total direct effect between the target object and the model prediction, with the goal of disentangling the positive effect from the negative side in a \textit{counterfactual} world. We prove that this goal can be achieved by strengthening the direct causal effect caused solely by the target object. Since the location of the target object is unknown, we propose patching-based inference to divide the original image into multiple patches and identify which patch contains the target object. Without the need for additional training, the method can be integrated into existing methods, enhancing their performance during inference in a flexible manner. Moreover, through the integration of patching into the training process, we introduce patching-based training to mitigate the feature distribution shift between the training and inference phases, resulting in further performance improvement. Experimental results on multiple benchmark datasets with diverse configurations verify that our method can achieve state-of-the-art performance.



  





\section{Related Work}
 
With the rapid development of deep models, MLC has recently attracted increasing attention. Existing solutions to the MLC problems can be roughly divided into three groups. The first kind of methods improved the traditional BCE loss to address the inherent positive-negative imbalance problem in the MLC tasks \cite{ridnik2021asymmetric}. 
The second kind focused on identifying regions of interest associated with semantic labels by utilizing the global average pooling (GAP) \cite{Verelst_2023_WACV} strategy or the attention technique \cite{lanchantin2021general,liu2021query2label,ridnik2023ml}. 
The last kind aimed at modeling label correlations, which have been regarded as fundamental information for MLC. The representative methods adopted a GCN \cite{chen2019multi} or developed a label masking strategy \cite{lanchantin2021general} to explore label correlations during training.
Recent works study how to improve multi-label classification performance with limited supervision, including semi-supervised multi-label learning \cite{cap2023,wang2020dual}, multi-label learning with partial labels \cite{durand2019learning,xie2022label}, and multi-label learning with noisy labels \cite{xie2022ccmn,li2022estimating}.

With the intensive consideration of correlations among variables, Peal's causal graphs \cite{10.1214/09-SS057,direct2001} and causal inference methods are widely used in machine learning. It provides solutions for achieving goals, \textit{e.g.}, improving model robustness by pursuing causal effects. Intervention-based and counterfactual-inference-based approaches were primarily utilized to achieve these goals. The intervention is mainly reflected in the control of confounding variables. For example, VC R-CNN \cite{wang2020visual} applied causal intervention to unsupervised regional feature learning, aiming to introduce a kind of ``visual common sense" into model design. CONTA \cite{zhang2020causal} attributed the cause of ambiguous pseudo-mask boundaries to the confusion context, and then removed the confusion bias via the backdoor criterion for weakly supervised semantic segmentation. CaaM \cite{wang2021causal}, as a novel causal attention module, self-annotated confounders in an unsupervised method and decoupled confounders and mediators via adversarial training. On the other hand, counterfactual-based techniques have often been applied to control or analyze causal relationships for tasks of interest. For example, while momentum may inadvertently bias tail predictions toward head predictions when training on long-tailed data, it also enhances representation learning and head predictions through an induced mediator \cite{tang2020long}. This paradoxical effect of momentum could be effectively disentangled by optimizing the total direct effect (TDE). 

\begin{figure*}[!tb]
	\centering
	\subfigure[Conditional True Positive Ratio]
	{
		\includegraphics[width=0.45\textwidth]{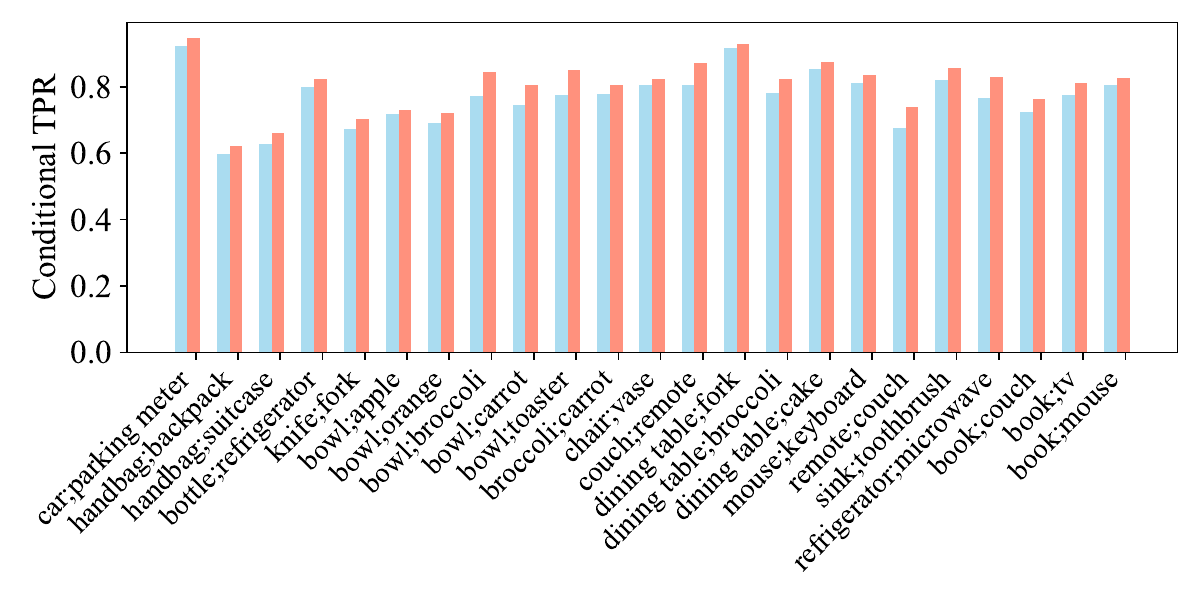}
		\label{fig:ctpr}
	}
	\subfigure[Conditional False Positive Ratio]
	{
		\includegraphics[width=0.45\textwidth]{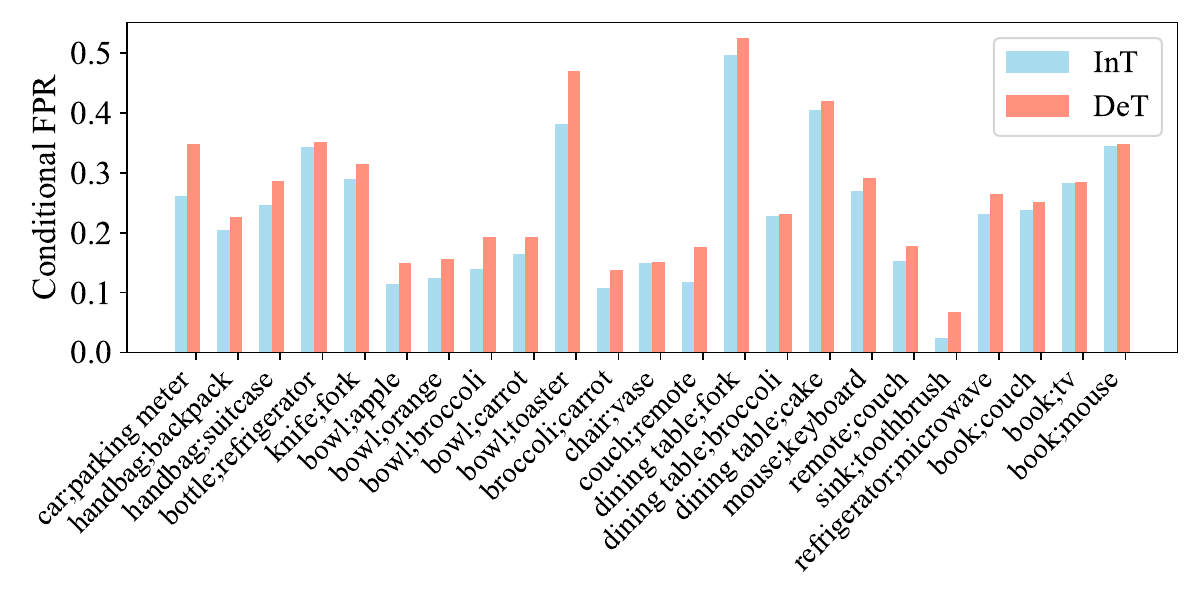}
		\label{fig:cfpr}
	}
	
	\caption{Comparison results between InT (each class has a visual backbone) and DeT (all classes share a visual backbone) in terms of conditional TPR (a) and FPR (b) for given class pairs on MS-COCO. DeT achieves greater conditional TPR and FPR than InT, meaning the co-occurrence relationships can be encoded by feature representations, exerting both positive and negative effects on model predictions.}
	\label{fig:tp_fp}
\end{figure*}

\begin{figure}[!t]
	\centering
	\includegraphics[width=0.45\textwidth]{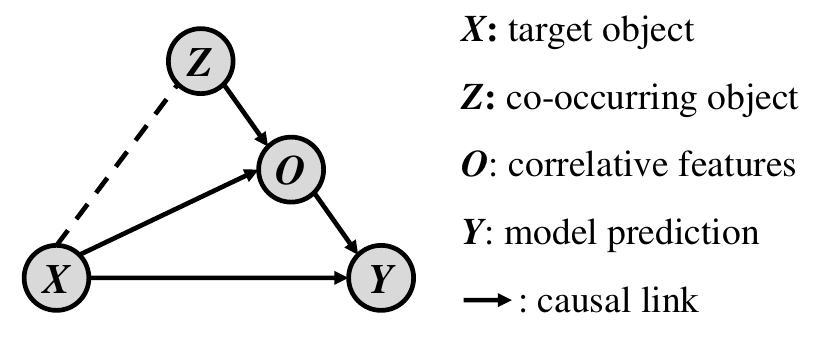}
	\caption{An illustration of the causal graph. The dashed line indicates these two objects co-occurring in an image.}
	\label{fig:cg}
\end{figure}

\section{Preliminaries}

In this section, we first provide necessary notations; and then we derive our results on the influence of co-occurrence for MLC.

\subsection{Notations}

Let $\i\in\mathcal{I}$ be an instance and $\y\subseteq\mathcal{Y}$ be its corresponding labels, where $\mathcal{I}=\mathbb{R}^d$ is the instance space and $\mathcal{Y}=\{0,1\}^q$ is the label space with $q$ possible class labels. Here, $y_k=1$ indicates that the $k$-th label is relevant to the instance; otherwise, $y_k=0$. Let $\mathcal{D}=\{(\i_j,\y_j)\}_{j=1}^n$ be the given training set with $n$ examples. Our goal is to train a deep neural network (DNN) $f(\cdot)$ based on the training set $\mathcal{D}$ that can output accurate predictions for unseen instances. We use $f(\i)$ to denote the predicted probabilities over all classes and $f_k(\i)$ to denote the predicted probability of the $k$-th class for instance $\i$. We use $[q]$ to denote the integer set $\{1,\cdots,q\}$.

\subsection{Co-Occurrence: A Double-Edged Sword for MLC}

We first provide a causal perspective to reveal that co-occurrence has positive and negative effects on MLC. As shown in Figure \ref{fig:cg}, we construct a causal graph to systematically study how co-occurrence relationships affect predictions. Without loss of generality, we assume that there are four variables in the causal graph: image content consisting of target object $X$ and its co-occurring object $Z$, correlative features $O$, and model prediction $Y$. Note that i) we do not know the specific locations of $X$ and $Z$ in the image; ii) the number of co-occurring objects can be more than one. In the causal graph, each causal link indicates causalities between two nodes, \textit{e.g.}, $X\rightarrow Y$ means that the prediction $Y$ is generated based on the features of $X$.

To disclose the practical rationality of our causal graph, let us first take a brief look at MLC in the context of deep learning. To train a DNN, the most commonly used MLC loss function is the BCE loss or its improved variant ASL loss \cite{ridnik2021asymmetric}, which decomposes the original task into multiple binary classification problems. For simplicity, we provide the definition of the BCE loss as
\begin{equation}\label{eq:bce}
\ell(f(\i),\y)=-\sum\limits_{k=1}^q y_k\log(f_k(\i))+(1-y_k)\log(1-f_k(\i)).
\end{equation}
Although the loss function does not explicitly consider label correlations, our interesting finding is that the model would inherently capture co-occurrence relationships via feature representation learning. We perform experiments on the benchmark dataset MS-COCO \cite{lin2014microsoft} to show this interesting finding. Specifically, we use two strategies to train the models, dependent training (DeT), which jointly trains a DNN for all classes together, \textit{i.e.}, the visual feature extractor is shared among all classes, and independent training (InT), which trains a DNN for each individual class, \textit{i.e.}, each class has its own feature extractor. Figure \ref{fig:ctpr} shows conditional true positive ratio (TPR) for a specific class given the presence of its co-occurring class, \textit{e.g.}, {\ttfamily{\{car;parking meter\}}} represents the proportion of images containing both objects {\ttfamily{car}} and {\ttfamily{parking meter}}, in which the model predicts the presence of the object {\ttfamily{car}}. From the figure, it can be observed that DeT achieves better performance than InT on given class pairs. This phenomenon goes against the intuition that InT should have outperformed DeT, since it trains an individual model for each class using all training examples, owning stronger learning capacity. The only difference between DeT and InT lies in the fact that the former utilizes a model to fit all relevant labels within an image, allowing for automatic capture of co-occurrence information. This results in better model performance on co-occurring classes. These results disclose that even without explicitly integrating co-occurrence relationships, DeT can capture correlative patterns, thereby enhancing the model's recognition ability.

This phenomenon can be explained by the link $(X,Z)\rightarrow O$, signifying that the target object and its co-occurring object jointly contribute to the correlative features $O$. Unlike InT, which derives the prediction $Y$ solely through the path $X\rightarrow Y$, DeT has an additional path $(X,Z)\rightarrow O\rightarrow Y$. This allows the model to enhance its performance on co-occurring objects by leveraging correlative patterns, as shown in Figure \ref{fig:ctpr}.


However, due to the powerful capacity of DNNs, they often tend to overfit to co-occurrence relationships. This can lead to incorrect predictions, indicating the presence of a particular object even when the image contains only its co-occurrences. Figure \ref{fig:cfpr} provides an empirical validation, showing the conditional false positive ratio (FPR) with respect to the same class pairs in Figure \ref{fig:ctpr}. From the figure, we can see that DeT achieves significantly larger conditional FPR than InT, which discloses that DeT suffers from severe overfitting to correlative features in the classes where it can effectively capture co-occurrence information.

This phenomenon can also be explained using our causal graph. Imagine a scenario where the object $X$ is not present in an image, but there is a co-occurring object $Z$. During the inference process, the model can still predict $Y$ for $X$ through the path $Z\rightarrow O\rightarrow Y$. The negative impact of co-occurrence relationships causes the model to make mistakes by predicting false positive labels. In one word, co-occurrence resembles a double-edged sword for MLC. On the positive side, it enhances the model's recognition ability by capturing co-occurrence information; on the negative side, it induces overfitting to correlative features, resulting in incorrect predictions.


\section{The Proposed Method}

To mitigate the negative side of mediated effect, a commonly used strategy is total direct effect (TDE) \cite{direct2001,vanderweele2013three}, designed to strengthen the direct causal effect along $X\rightarrow Y$. In the context of MLC, for a given image $\i$, we define $\text{TDE}_k(\i)$ as the probability that the target object $\x$ in the image belongs to class $k$.
\begin{equation}\label{eq:tde}
\begin{aligned}
\text{TDE}_k(\i)&=P(Y_{\o}^k=1|X=\x, Z=\z)\\
	&\qquad-P(Y_{\o}^k=1|X=\x_0, Z=\z),
\end{aligned}
\end{equation}
where $X=\x_0$ represents the operation of masking the features of target object $\x$. The subscript $\o$ represents that the mediator $O$ takes the effect of the correlative features $\o$ on the prediction. The goal of TDE is to preserve the positive side via the mediation path while mitigating the negative effect caused by the mediator. Specifically, the first term in Eq.\eqref{eq:tde} represents the prediction by incorporating the co-occurrence information $\o$ caused by $\x$ and $\z$; while the second term denotes the prediction when the target object $\x$ is masked (denoted by $\x_0$) but the co-occurrence information is still $\o$ caused by the co-occurring object $\z$. It is noteworthy that the second term actually measures the bias of overfitting to the co-occurrence relationship between $X$ and $Z$, \textit{i.e.}, the model still predicts target label $Y^k$ as positive in the presence of the co-occurring object $Z$, despite the absence of the target object $X$. The subtraction operator exactly removes the negative influence of the mediated effect.

\begin{figure}[!t]
	\centering
	\includegraphics[width=0.45\textwidth]{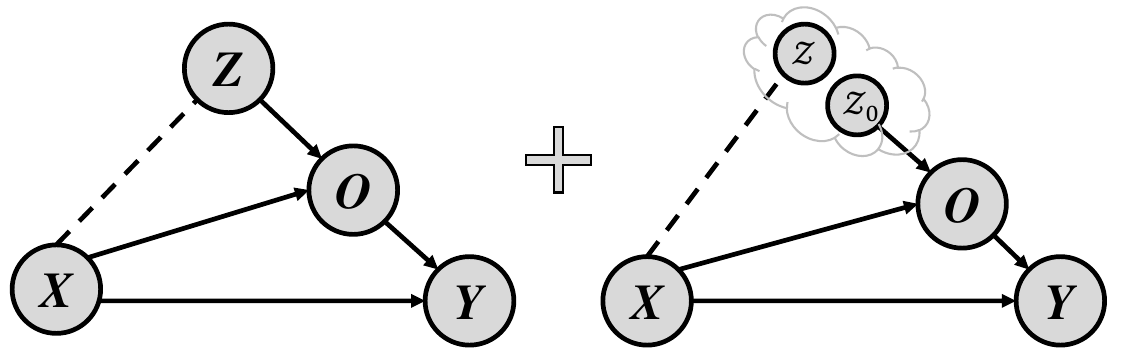}
	\caption{An illustration of TDE inference for MLC. }
	\label{fig:tde}
\end{figure}

During inference, it is straightforward to obtain the first probability of TDE by training a model with Eq.~\eqref{eq:bce}. Unfortunately, due to the unknown location of the target object, it is hard to obtain the second probability by masking the features of $\x$. Considering that the subtraction operator often introduces the issue of negative values, we transform the second probability into its complementary form, \textit{i.e.}, the probability for the target object $\x$ when masking its co-occurring object $\z$. Then, using this complementary probability, as illustrated in Figure \ref{fig:tde}, we rewrite TDE as the following equivalent form \footnote{The details can be found in Appendix \ref{app:tde}. }
\begin{equation}\label{eq:tde2}
	\begin{aligned}
		\text{TDE}_k(\i)&=P(Y_{\o}^k=1|X=\x, Z=\z)\\
		&\qquad+\lambda P(Y_{\o}^k=1|X=\x, Z=\z_0),
	\end{aligned}
\end{equation}
where $Z=\z_0$ represents the operation of masking the features of the co-occurring object $\z$, and $\lambda$ is a dataset-dependent parameter. The intuition behind this transformation is that it transfers the mitigation of the negative mediated effect caused only by the co-occurring objects $\z$ into the enhancement of the direct causal effect caused only by the target object $\x$. Compared to InT, TDE leverages the co-occurrence information by considering the mediated effect along $(X,Y)\rightarrow O\rightarrow Y$ (the first term in Eq.~\eqref{eq:tde2}), significantly enhancing model performance; compared to DeT, TDE captures discriminative patterns by enhancing the direct effect along $X\rightarrow Y$ (the second term in Eq.~\eqref{eq:tde2}), which prevents the model from overfitting to correlative features. 



\paragraph{Patching-Based Inference} To obtain the predicted probability $P(Y^k_{\o}=1|X=\x,Z=\z_0)$, we propose a patching-based inference (PAT-I) method. The main idea involves partitioning the original image into multiple patches and subsequently determining which patch contains the target object. Then, the prediction for the identified patch can be considered as an approximation of the prediction for the target object. Specifically, during inference, for each image $\i$, we crop it into $m$ equally-sized patches $\{\u_j\}_{j=1}^m$. By feeding the original image and the patches into the trained model $f(\cdot)$, we obtain the predictions $\p=[p_1,...,p_q]$ and $\q_j=[q_{j1},...,q_{jq}]$, where $p_k=f_k(\i)$ and $q_{jk}=f_k(\u_j)$ denote the predicted logits of the class $k$ for the image $\i$ and patch $\u_j$ respectively. 

Given the probability that the target object may be present in multiple patches, rather than selecting a particular patch, we assign a weight to each patch based on its predictions
\begin{equation}\label{eq:weight}
\forall j\in[m], \w_j=\frac{\exp(\q_j/\tau)}{\sum_{l=1}^m\exp(\q_l/\tau)},
\end{equation}
where $\tau$ is a temperature parameter that controls the degree of emphasis on a particular patch. Then, the predicted logits for different objects can be obtained by aggregating the logits $\q_j$ in a spatially-weighted manner, $\q=\sum\limits_{j=1}^m\w_j\cdot\q_j$. Finally, according to Eq.~\eqref{eq:tde2}, we obtain the model prediction for $\x$ of class $k$ as $\text{TDE}_k(\i)=\delta(p_k+\lambda\cdot q_k$), where $\delta(\cdot)$ is the \textit{Sigmoid} function. In our experiments, we observed that setting $\lambda=1$ achieves favorable performance. The reason for performing above operations on logits instead of probabilities is that we found applying the \textit{Sigmoid} function to logits to convert them into probabilities compresses the distinctive information, which in turn hinders the effectiveness of our method.
Generally, PAT-I has two main advantages: i) simplicity - it can be incorporated directly into the inference process to enhance model performance without any additional training; ii) flexibility - it can be seamlessly integrated into any trained model.


\paragraph{Patching-Based Training} Although PAT-I can improve model performance when integrated into the inference process, it often encounters the challenge of feature distribution shift between the original images and patches. This is because the model is trained only on original images that exhibit a significant distribution shift with patches. This may cause the model to produce unconfident predictions, resulting in suboptimal performance. To address this problem, we propose patching-based training (PAT-T) to utilize both images and patches for model training, which aims to alleviate the distribution shift between the training and inference phases.

Specifically, we decompose the trained model $f(\cdot)$ into two components $f(\cdot)=h\circ g(\cdot)$, where $g(\cdot)$ is a backbone and $h(\cdot)$ is a classification head. During training, for each image $\i$, we obtain the predictions $\p$ and $\q_j$ in the same way. In practice, to further improve the model performance, we make two modifications. Firstly, considering that the classification head is sensitive to two different classification tasks on images and patches, we adopt two classifier heads $h_{\phi}$ and $h_{\psi}$ to make predictions on images and patches independently, \textit{i.e.}, $\p=h_{\phi}\circ g(\i)$ and $\q_j=h_{\psi}\circ g(\u_j)$, where $\phi$ and $\psi$ denote two sets of parameters. Secondly, we also use another head $h_{\theta}$ to obtain the predicted logits $\q^w_j=h_{\theta}\circ g(\u_j)$, which are used to estimate weights $\w_j$ according to Eq.~\eqref{eq:weight}. By decoupling the dependence between weight estimation and logits prediction, it enhances model performance by avoiding the accumulation of prediction errors on a single head. 



Given the predictions $\p$ and $\q$ on the image and patches, we define the classification loss of PAT-T as 
\begin{equation}\label{eq:final}
\mathcal{L}=\ell(\delta(\p),\y)+\ell(\delta(\q),\y),
\end{equation}
where $\ell(\cdot,\cdot)$ can be BCE or ASL loss. In our experiments, we use ASL loss because it often achieves better performance than BCE loss. 






\section{Experiments}


\subsection{Experimental Settings}

\begin{table*}[!t]
	\centering
	\caption{Comparison results on MS-COCO between our method and state-of-the-art methods. The backbones denoted with 21k and OI are pretrained on the ImageNet-21k and Open Image datasets, respectively. The best performance is highlighted in bold. All metrics are in \%.}
	\begin{tabular}{l | c | c | c | c | c | c | c | c | c}
		\midrule\midrule
		Method  & Backbone        & Resolution & mAP   & CP     & CR    & CF1   & OP & OR & OF1 \\
		\midrule
		ResNet-101 \cite{he2016deep} & ResNet101  & $224\times 224$ & 78.3 & 80.2 & 66.7 & 72.8 & 83.9 & 70.8 & 76.8               \\
		ML-GCN \cite{chen2019multi}     & ResNet101  & $448\times 448$ & 83.0 & 85.1 & 72.0 & 78.0 & 85.8 & 75.4 & 80.3               \\
		KSSNet \cite{wang2020multi}     & ResNet101  & $448\times 448$ & 83.7 & 84.6 & 73.2 & 77.2 & 87.8 & 76.2 & 81.5               \\
		MS-CMA \cite{you2020cross}     & ResNet101  & $448\times 448$ & 83.8 & 82.9 & 74.4 & 78.4 & 84.4 & 77.9 & 81.0               \\
		MCAR \cite{gao2021learning}   & ResNet101  & $448\times 448$ & 83.8 & 85.0 & 72.1 & 78.0 & 88.0 & 73.9 & 80.3               \\
		CPSD \cite{xu2022boosting}      & ResNet101  & $448\times 448$ &  83.1 & 83.5 & 73.6 & 78.2 & 84.8 & 77.3 & 80.9              \\
		IDA \cite{liu2023causality}        & ResNet101  & $448\times 448$ &  84.8 & -    & -    & 78.7 & -    & -    & 80.9     \\
		\midrule 
		PAT-T      & ResNet101  & $448\times 448$ & 85.0 & 87.6 & 74.4 & 80.5 & 88.4 &	77.9 & 82.8 \\ 
		\midrule
		ASL \cite{ridnik2021asymmetric}       & TResNetL       & $448\times 448$ & 86.6 & 87.2 & 76.4 & 81.4 & 88.2 & 79.2 & 81.8     \\
		CPSD  \cite{xu2022boosting}      & TResNetL       & $448\times 448$ & 87.3 & 85.5 & 78.9 & 82.1 & 85.7 & 81.5 & 83.7 \\
		TResNetL \cite{ridnik2021imagenetk}  & TResNetL (21k) & $448\times 448$ & 88.4 & -    &  -   & -    & -    &  -   & - \\
		\midrule
		PAT-T      & TResNetL       & $448\times 448$ & 88.7 & 89.0 & 78.8 & 83.6 &	89.3 &	81.3 & 85.1 
		 \\
		PAT-T      & TResNetL (21k) & $448\times 448$ & 90.6 & 90.6 & 81.0 & 85.5 &	90.9 &	82.9 & 86.7 
		 \\
		\midrule
		Q2L-TResL \cite{liu2021query2label} & TResNetL       & $448\times 448$ & 87.3 & 87.6 & 76.5 & 81.6 & 88.4 & 78.5 & 83.1 \\
		Q2L-TResL  & TResNet (21k)  & $448\times 448$ & 89.2 & 86.3 & 81.4 & 83.8 & 86.5 & 83.3 & 84.9 \\
		\midrule
		PAT-T + Q2L-TResL  & TResNetL       & $448\times 448$ & 89.5 & 89.5 & 79.3 &	84.1 & 90.0 & 81.0 & 85.3  \\
		PAT-T + Q2L-TResL  & TResNetL (21k) & $448\times 448$ & 91.0 & 89.1 & 82.7 &	85.8 &	89.8& 	83.9 &	86.8 
		 \\
		\midrule
		ML-Dec \cite{Ridnik_2023_WACV}  & TResNetL (OI)   & $448\times 448$ & 90.0 & -    & -    & -     & -     &   -   & -    \\ 
		Q2L-TResL       & TResNetL (OI)   & $448\times 448$ & 90.1 & 83.6 &	84.2 & 	83.9 & 	85.0 & 	85.9 &	85.4 
		  \\ 
		Q2L-SwinL       & Swin-L(21k)     & $384\times 384$ & 90.5 & 89.4 & 81.7 & 85.4 & 89.8 & 83.2 & 86.4 \\ 
		Q2L-CvT         & CvT-w24(21k)    & $448\times 448$ & 91.3 & 88.8 & 83.2 & 85.9 & 89.2 & 84.6 & 86.8 \\
		\midrule
		PAT-T + ML-Dec& TResNetL (OI)   & $448\times 448$ & 91.4 & 90.1 & 82.9 & \textbf{86.4}  & 90.5 & 84.8 & \textbf{87.6}  \\
		PAT-T + Q2L-TResL & TResNetL (OI)   & $448\times 448$ & \textbf{91.7} & 89.7 & 83.4 & \textbf{86.4} & 90.3 & 84.9 & 87.5   \\
		\midrule\midrule
	\end{tabular}
	\label{tb:coco}
\end{table*}

\paragraph{Dataset} To evaluate the performance of the proposed method, we conduct experiments on three benchmark datasets, including MS-COCO 2014 \footnote{\url{https://cocodataset.org}} \cite{lin2014microsoft}, Pascal VOC 2007 \footnote{\url{http://host.robots.ox.ac.uk/pascal/VOC/}} \cite{everingham2010pascal}, and Visual Genome \footnote{\url{https://homes.cs.washington.edu/~ranjay/visualgenome/index.html}} \cite{krishna2017visual}. MS-COCO contains 82,081 training images and 40,137 validation images for 80 classes, with an average of 2.9 labels per image. VOC 2007 consists of the \textit{train-val} set with 5,011 images and \textit{test} set with 4,952 images. Following previous works \cite{ridnik2021asymmetric, liu2021query2label}, we use \textit{train-val} set for training and \textit{test} set for testing. Visual Genome is a dataset that contains 108,249 images and covers 80,138 categories. Similar to the previous work \cite{chen2022structured}, considering that most categories have very few examples, we conduct preprocessing on this dataset. The main steps include: i) merge categories: combine categories that share the same meaning; ii) filter by image count: exclude categories with fewer than 500 images. Finally, we obtain a dataset named VG-256 that contains 256 classes and 108,249 images. We randomly divide 70\% for training and 30\% for testing.

\paragraph{Evaluation Metric} Following previous works \cite{chen2019multi, ridnik2021asymmetric}, we utilize the average precision (AP) for individual categories and the mean average precision (mAP) across all categories as our evaluation metrics. To provide a comprehensive assessment of model performance, we additionally showcase overall precision (OP), recall (OR), F1-measure (OF1), as well as per-category precision (CP), recall (CR), and F1-measure (CF1) for detailed comparisons. 


\paragraph{Implementation Details}

Following previous works \cite{ridnik2021asymmetric, liu2021query2label}, we choose ResNet101 and TResNetL \cite{ridnik2021tresnet} pre-trained on ImageNet \cite{deng2009imagenet} or Open Image \cite{papadopoulos2017extreme} as the backbone for our method. We employ Adam \cite{kingma2014adam} optimizer and one-cycle policy scheduler to train the model with maximal learning rate of 0.0001. Furthermore, we perform exponential moving average (EMA) \cite{tarvainen2017mean} for the model parameters with a decay of 0.9997. We use the ASL loss as the base loss function since it shows superiority to BCE loss. We perform all experiments on GeForce RTX 3090 GPUs. The random seed is set to 1 for all experiments.

\section{Comparison with State-of-the-art Methods}

\begin{table*}[!t]
	\setlength{\tabcolsep}{0.1em}
	\small
	\centering
	\caption{Comparison results on VOC 2007 between our method and state-of-the-art methods in terms of AP (\%) and mAP (\%). All results are reported at resolution $448\times 448$ except for the ADD-GCN and SSGRL, whose resolutions are $576\times576$. The best performance is highlighted in bold.}
	\begin{tabular}{|c | c c  c  c  c  c  c  c  c c  c  c  c  c  c  c  c  c  c c|c|}
		\hline
		Methods &aero&bike&bird&boat&bottle&bus&car&cat&chair&cow&table&dog&horse&mbike&person&plant&sheep&sofa&train&tv&mAP\\
		\hline
		CNN-RNN &96.7&83.1&94.2&92.8&61.2  &82.1 & 89.1 & 94.2 & 64.2 & 83.6 & 70.0& 92.4& 91.7& 84.2& 93.7& 59.8& 93.2& 75.3& \textbf{99.7}& 78.6& 84.0\\
		VGG+SVM & 98.9& 95.0& 96.8& 95.4& 69.7& 90.4 &93.5& 96.0 &74.2& 86.6& 87.8& 96.0& 96.3& 93.1& 97.2& 70.0& 92.1& 80.3 &98.1& 87.0 &89.7\\
		Fev+Lv& 97.9 &97.0& 96.6& 94.6& 73.6& 93.9& 96.5& 95.5& 73.7& 90.3& 82.8 &95.4 &97.7 &95.9& 98.6& 77.6& 88.7& 78.0 &98.3& 89.0& 90.6\\
		HCP & 98.6 & 97.1 & 98.0 & 95.6& 75.3& 94.7& 95.8& 97.3& 73.1& 90.2& 80.0& 97.3& 96.1& 94.9& 96.3& 78.3& 94.7& 76.2& 97.9& 91.5& 90.9\\	
		RDAL& 98.6& 97.4& 96.3& 96.2& 75.2& 92.4& 96.5& 97.1& 76.5& 92.0& 87.7& 96.8& 97.5& 93.8& 98.5& 81.6& 93.7& 82.8& 98.6& 89.3& 91.9\\
		RARL& 98.6& 97.1& 97.1& 95.5& 75.6& 92.8& 96.8& 97.3& 78.3& 92.2& 87.6& 96.9& 96.5& 93.6& 98.5& 81.6& 93.1& 83.2& 98.5& 89.3& 92.0\\
		SSGRL & 99.7& 98.4& 98.0& 97.6& 85.7& 96.2& 98.2& 98.8& 82.0& 98.1& 89.7& 98.8& 98.7& 97.0& 99.0& 86.9& 98.1& 85.8& 99.0& 93.7& 95.0\\
		MCAR& 99.7& \textbf{99.0}& 98.5& 98.2& 85.4& 96.9& 97.4& 98.9& 83.7& 95.5& 88.8& 99.1& 98.2& 95.1& 99.1& 84.8& 97.1& 87.8& 98.3& 94.8& 94.8\\
		ASL(TResL)& 99.9 & 98.4 & 98.9& 98.7& 86.8& 98.2& 98.7& 98.5& 83.1& 98.3& 89.5& 98.8& 99.2& 98.6& \textbf{99.3}& 89.5& \textbf{99.4}& 86.8& 99.6& 95.2& 95.8\\
		ADD-GCN & 99.8 & \textbf{99.0}& 98.4& \textbf{99.0}& 86.7& 98.1& 98.5& 98.3& \textbf{85.8}& 98.3& 88.9& 98.8& 99.0& 97.4& 99.2& 88.3& 98.7& \textbf{90.7}& 99.5& \textbf{97.0} & 96.0\\
		Q2L-TResL& 99.9& 98.9& 99.0& 98.4& 87.7& 98.6& \textbf{98.8}& \textbf{99.1}& 84.5& 98.3& 89.2& 99.2& 99.2& \textbf{99.2}& \textbf{99.3}& \textbf{90.2}& 98.8& 88.3& 99.5& 95.5& 96.1\\
		\hline
		PAT-T + TResL& \textbf{100} & 98.8& \textbf{99.3}& 98.6& \textbf{88.8}& \textbf{99.0} & \textbf{98.8} & \textbf{99.1}&  84.5&  \textbf{98.9}& \textbf{90.1} &  \textbf{99.5}& \textbf{99.4} &  98.6&  \textbf{99.3}&  89.2& \textbf{99.4} &  86.3& \textbf{99.7} &  95.8& \textbf{96.2}\\
		\hline
	\end{tabular}
	\label{tb:voc}
\end{table*}

\begin{table*}[!t]
	\centering
	\caption{Comparison results on VG-256 between our method and state-of-the-art methods. All results are reported at resolution $448\times 448$. The best performance for each backbone is highlighted in bold.
	}
	\begin{tabular}{c | c |  c | c | c | c | c | c | c}
		\midrule\midrule
		Method  & Backbone & mAP   & CP     & CR    & CF1   & OP & OR & OF1 \\
		\midrule
		ResNet101    & ResNet101      & 47.7  &	51.1 & 48.2 & 49.6 & 63.3 &	64.0 &	63.6 \\
		PAT-I        & ResNet101      & 48.5  &	53.9 & 46.3 & 49.8 & 64.0 &	64.8 &	64.4  \\
		PAT-T        & ResNet101      & \textbf{49.9}  & 52.0 & 51.7 & \textbf{51.9} & 63.5 &	66.0 &	\textbf{64.7}  \\
		\midrule
		TResNetL     & TResNetL (21k) & 53.9  & 57.2 & 53.4 & 55.2 & 66.7 &	67.7 & 67.2  \\
		PAT-I        & TResNetL (21k) & 54.8  &	58.7 & 52.5 & 55.4 & 66.1 &	69.1 &	67.6 
		 \\
		PAT-T        & TResNetL (21k) & \textbf{56.2}  & 58.0 & 56.0 & \textbf{57.0} & 67.1 &	69.3 &	\textbf{68.2} 	\\
		\midrule
		Q2L-TResL    & TResNetL (OI) & 57.7 & 58.7 & 58.6 & 58.7 &	67.4 & 70.6 & 69.0 \\
		PAT-I+Q2L-TResL & TResNetL (OI) & 58.0 & 58.8 & 58.3 &	58.6 &	66.1 &	72.0 &	68.9 \\
		PAT-T+Q2L-TResL & TResNetL (OI) & \textbf{59.5} & 59.6 & 59.9 &	\textbf{59.7} &	67.9 &	71.7 &	\textbf{69.8} \\
		\midrule\midrule
	\end{tabular}
	\label{tb:vg}
\end{table*}

\subsection{Performance on MS-COCO}

Table \ref{tb:coco} reports the comparison results between our method PAT-T and state-of-the-art methods on MS-COCO. Due to the page limit, the results of PAT-I can be found in Appendix \ref{sec:moreres}. We can see that our method achieves better performance than the comparing methods with the same configuration in terms of mAP, CF1 and OF1, which are the most important metrics, as other metrics are sensitive to the threshold. In particular, under the same configuration, \textit{i.e.}, the same pre-trained back and resolution, PAT-T improves upon the performance of ASL by 2.1\%, Q2L by 2.2\%, ML-Decoder by 1.4\%, respectively. Moreover, the optimal mAP score (91.7\%) achieved by PAT-T is better than the current state-of-the-art mAP score (91.3\%) achieved by Q2L-CvT. The backbone TResNet (54.7M) used by our method has a significantly smaller number of model parameters compared to CvT-w24 (277M) used by Q2L-CvT. These results convincingly show that PAT can achieve state-of-the-art performance based on a far smaller backbone.

\subsection{Performance on VOC 2007}

Table \ref{tb:voc} reports the comparison results on VOC 2007. To make a fair comparison, following the previous works \cite{ridnik2021asymmetric,liu2021query2label}, we use TResNet pre-trained on ImageNet-1K as the backbone. The results based on a stronger backbone can be found in Appendix \ref{sec:moreres}. Our method achieves the best performance when compared to state-of-the-art methods. Moreover, our method achieves the optimal performance in the majority of classes. It seems that our method does not yield as significant improvement as it did on MS-COCO. Except for the small number of images and relatively high performance achieved on VOC 2007, one possible reason may be that the co-occurrence relationships in the dataset are not as strong as in MS-COCO, as each example has only 1.5 labels on average. This makes models less prone to overfitting to label correlations.


\subsection{Performance on VG-256}

Table \ref{tb:vg} reports the comparison results between our methods PAT-I, PAT-T and the comparing methods on VG-256. We run all methods in the resolution of $448\times 448$. From the table, we can see that PAT-I can improve the model performance without any additional training cost. It enhances the positive impact while alleviating the negative impact of the mediated effect during the inference phase. Furthermore, by incorporating this idea into training, PAT-T achieves significant performance improvements and obtains the best performance. These results strongly verify the effectiveness of the proposed method.

\begin{figure*}[!tb]
	\centering
	\includegraphics[width=0.85\textwidth]{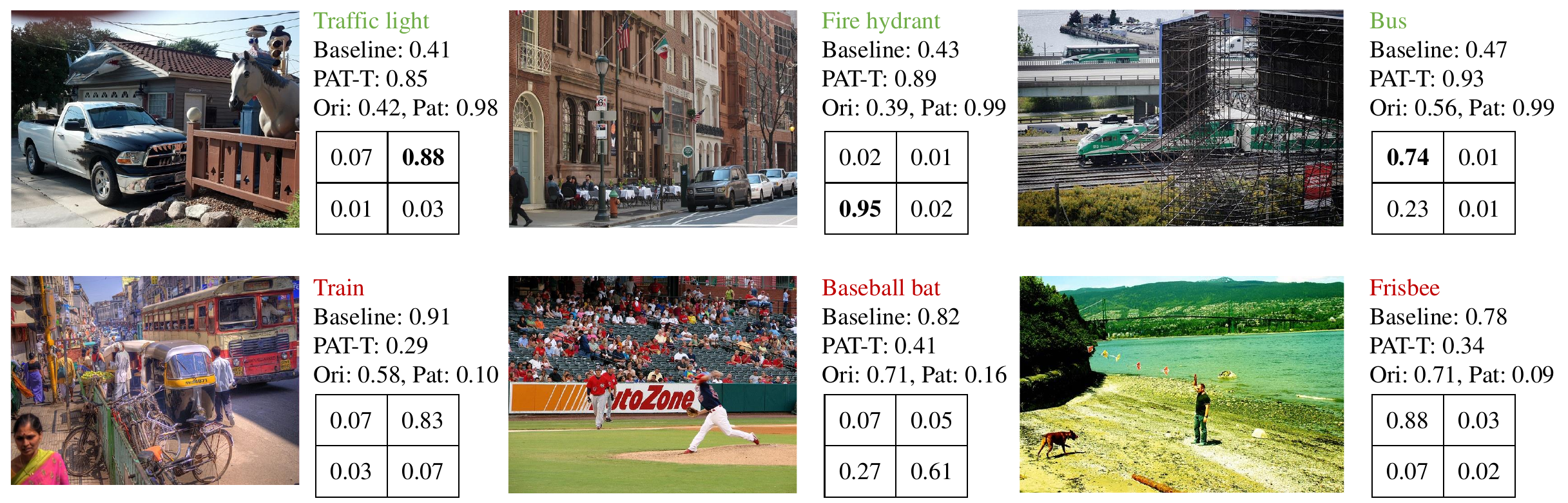}
	\caption{Visualization of model predictions on MS-COCO. Ori denotes the probability predicted by PAT-T model on the original image; Pat denotes the probability predicted by PAT-T model on patches. The weights of each patch are reported in the four-grid.}
	\label{fig:case}
\end{figure*}

\begin{figure}[!tb]
	\centering
	
	\subfigure
	{
		\includegraphics[width=0.4\textwidth]{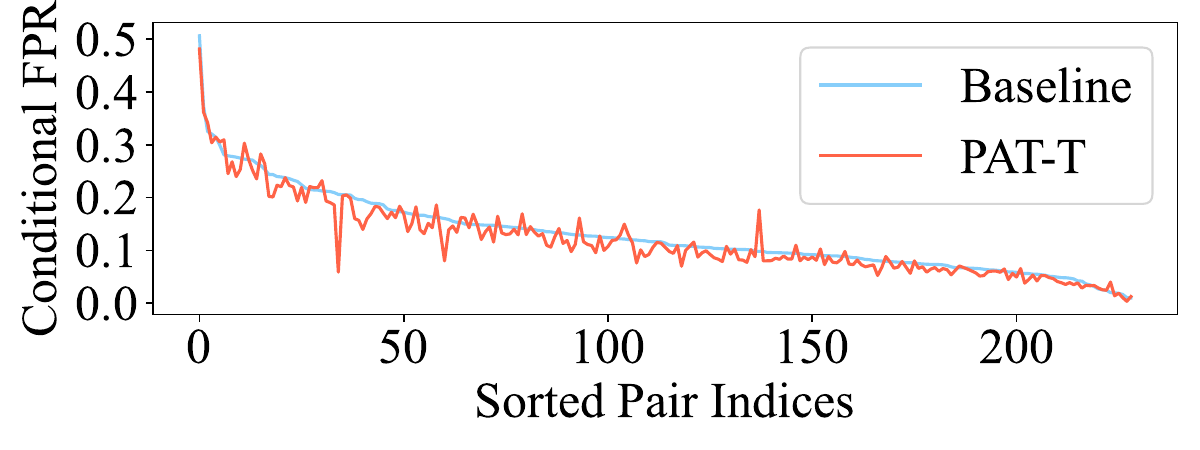}
	}\vspace{-0.5cm}
	
	\subfigure
	{
		\includegraphics[width=0.4\textwidth]{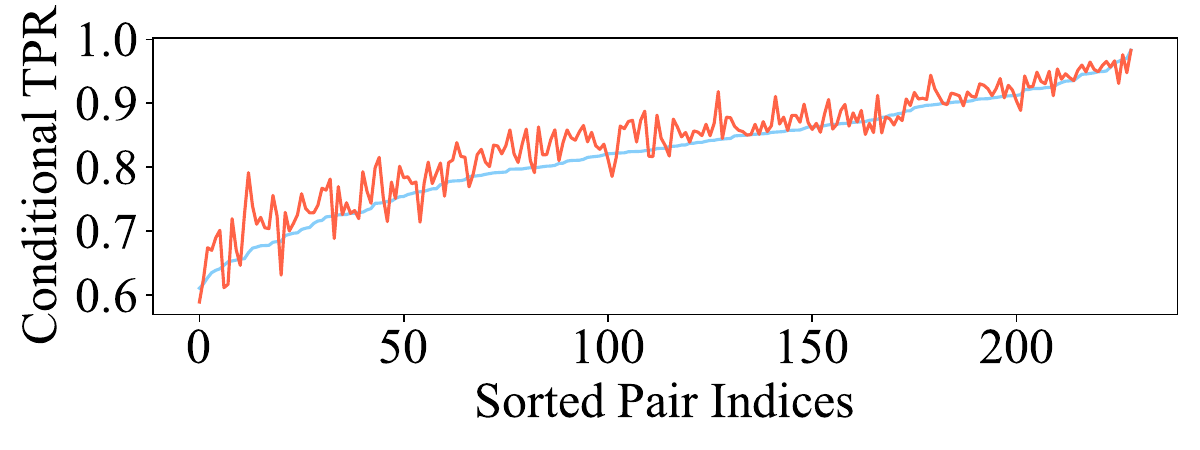}
	}
	\caption{Conditional FPR and conditional TPR of the label pairs with co-occurrence probabilities larger than 0.2 on MS-COCO. The pair indices are sorted according to the performance of Baseline. PAT-T outperforms Baseline in around 76\% and 77\% percentage of label pairs in terms of CFPR and CTPR.}
	\label{fig:curve}
	\vspace{-0.5cm}
\end{figure}

\subsection{Ablation Studies}

To demonstrate the effectiveness of our method in preventing models from overfitting to co-occurrence relationships, Figure \ref{fig:curve} respectively illustrates the conditional FPR and conditional TPR of label pairs with a co-occurrence probability greater than 0.2. Baseline (ASL) and our PAT-T are trained with the same configuration (the backbone is TResNetL). In order to reduce the apparent fluctuations in the curve, we sort the label pairs according to the performance of Baseline. From the figure, we can see that the curve of PAT-T is generally below ASL, yielding that PAT-T can prevent the model from overfitting to co-occurrence, thereby making fewer mistakes in predicting false positive labels. Moreover, PAT-T also outperforms Baseline in terms of conditional TPR, indicating that PAT-T enhances the recognition capacity of the model significantly.

\subsection{Step-Wise Evaluation}

\begin{figure}[!tb]
	\centering
	\setcounter{subfigure}{0}
	\subfigure[Original Prediction $\p$ vs Patch Prediction $\q$]
	{
		\includegraphics[width=0.48\textwidth]{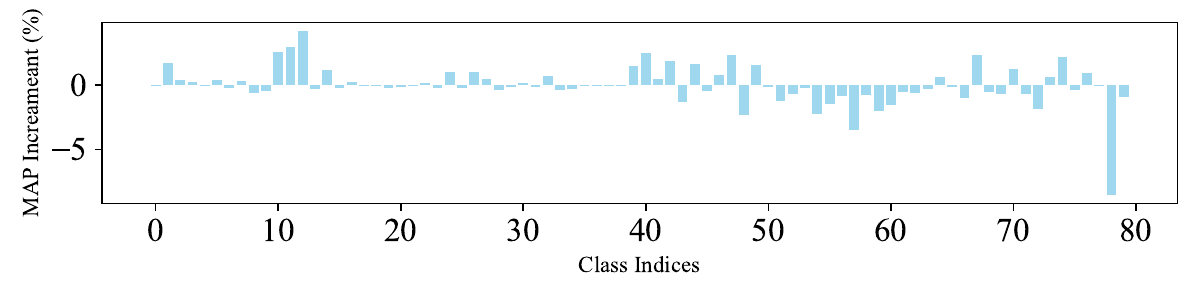}
	}\label{fig:bara}
	\vspace{-0.2cm}
	
	\subfigure[Optimum between $(\p,\q)$ vs TDE Prediction $\p+\lambda\cdot\q$]
	{
		\includegraphics[width=0.48\textwidth]{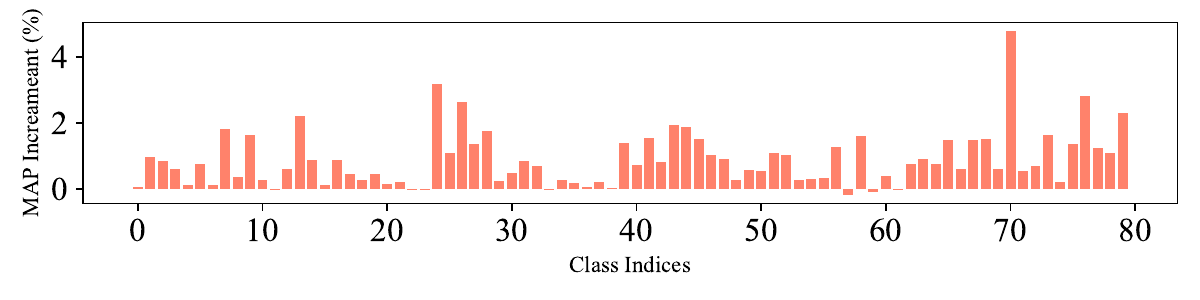}
	}\label{fig:barb}
	\caption{Corresponding to three output logits, original logits $\p$ on an image, patch logits $\q$ on patches, and TDE logits $\p+\lambda\cdot\q$, we illustrate the per-class average precision increments between performance of each two logits to show the effectiveness of TDE.}
	\label{fig:bar}
	\vspace{-0.5cm}
\end{figure}

We perform a step-wise evaluation on MS-COCO to have a better understanding of how TDE keeps the positive side while removing the negative side of the mediated effect. Figure \ref{fig:bar} (a) shows there is no significant difference between the performance of original logits $\p$ and patch logits $\q$. In some classes, the former performs better, while in other classes, the latter exhibits superior performance. This means that introducing the patching technique does not lead to an improvement in model performance. Figure \ref{fig:bar} (b) illustrates the improvements in performance of the proposed TDE logits on the basis of the optimum between the original logits $\p$ and patch logits $\q$. We can see TDE maintains positive effects in almost every class, achieving the retention of positive causal effect and the removal of negative one. 

\subsection{Case Studies}

To disclose the mechanism behind PAT-T, as shown in Figure \ref{fig:case}, we visualize some cases of predictions on MS-COCO. The first row reports the predictions of indistinguishable objects, \textit{e.g.}, obscured, tiny, or ambiguous objects, when their co-occurring counterparts are present in the image. For such objects, our PAT-T gives the probabilities that are significantly higher than Baseline. This is because by focusing on a single patch instead of a whole image, our method is able to identify these objects (which can be validated by the probabilities on patches denoted by \textit{Pat} and weights in the four-grid), thereby enhancing the direct causal effect caused by target objects. The second row reports the predictions of objects that are absent, yet their co-occurring counterparts are present in the image. We can see Baseline suffers from overfitting to the correlative features caused by co-occuring objects. For example, in the second image, Baseline predicts a relatively high probability for \textit{baseball bat} due to the presence of its co-occurring object \textit{baseball glove}. While PAT-T predicts a smaller probability (0.41), since it does not find the target object in any patch, which can be validated by the predicted probabilities on patches (0.16).

\begin{table}[!t]
	\centering
	\caption{Comparison results on MS-COCO between patch logits $\q$ and TDE logits $\p+\lambda\cdot\q$.}
	\begin{tabular}{c | c | c }
		\midrule\midrule
		Backbone  & Patch Logits  & TDE Logits \\
		\midrule
		TResNetL  & 87.1  &  88.7  \\
		TResNetL (21k)  & 88.8 &  89.2  \\
		TResNetL (OI)  & 89.5 &  90.8  \\
		\midrule
		Q2L-TResL  & 87.9 &  89.5  \\
		Q2L-TResL (21k)  & 89.1 &  91.0  \\
		Q2L-TResL (OI)  & 90.1 &  91.7  \\
		\midrule\midrule
	\end{tabular}
	\label{tb:scale}
\end{table}

\subsection{Study on Influence of Image Scale}

In the PAT method, we found that training a backbone with images and patches results in a significant decline in performance. This is due to their inconsistent sizes, \textit{i.e.}, the patches are only a quarter of the size of the image, severely disrupting feature representation learning. To deal with this problem, we first resize the patches to the size of the original image before inputting them into the model. Note that this operation does not incorporate additional information, because we first crop the resized image into patches and then resize them again. This operation only affects the scale of the targets. We perform experiments to verify that the main reason for the performance improvement is not due to the increase in object scale. Specifically, we train a model based on only resized patches using the loss function $\ell(\delta(\q),\y)$ in Eq. \eqref{eq:final}. Table \ref{tb:scale} reports mAP scores of patch logits and TDE logits. From the table, we can see that
when the representation capability of the backbone is relatively weak, patch logits performs slightly better than the image logits (baseline results in Table \ref{tb:coco} with corresponding backbones). As the representation capability of the backbone increases, this advantage disappears, and there is no significant difference in performance between patch logits and image logits. While our TDE logits still outperform patch logits with a large margin. These results convincingly validate that the effectiveness of our method stems from decoupling co-occurrence relationships using TDE.

\section{Conclusion}

The paper studies the problem of multi-label learning, which aims to improve model performance by leveraging label correlations. Previous studies demonstrate that the powerful fitting capacity of DNNs often leads to overfitting to co-occurrence relationships. We establish a causal reasoning framework that treats the correlative features caused by the target object and its co-occurring ones as a mediator, exhibiting both positive and negative impacts on model predictions. To keep the positive side while removing the negative impact, we propose to enhance the direct causal effect caused by the target object when masking its co-occurring objects. In implementation, we propose patching-based training to identify the pivot patch that contains the target object. Experimental results on multiple benchmark datasets validate the superiority of our method to state-of-the-art methods.

\section*{Impact Statement}

This paper presents work whose goal is to advance the field of Machine Learning. There are many potential societal consequences of our work, none which we feel must be specifically highlighted here.

\section*{Acknowledgements}

Sheng-Jun Huang was supported by Natural Science Foundation of Jiangsu Province of China (BK20222012, BK20211517), the National Key R\&D Program of China (2020AAA0107000), and NSFC (62222605). Masashi Sugiyama was supported by JST CREST Grant Number JPMJCR18A2.


\bibliography{ref}
\bibliographystyle{icml2024}

\newpage
\appendix
\onecolumn

\begin{table*}[!t]
	\setlength{\tabcolsep}{0.1em}
	\small
	\centering
	\caption{Comparison results on VOC 2007 between our method and state-of-the-art methods in terms of AP (\%) and mAP (\%). All results are reported at resolution $448\times 448$ except for the ADD-GCN and SSGRL, whose resolutions are $576\times576$. The best performance is highlighted in bold. The asterisk (*) indicates the results reproduced using the released codes, as they were not provided in the original papers.}
	\begin{tabular}{|c | c c  c  c  c  c  c  c  c c  c  c  c  c  c  c  c  c  c c|c|}
		\hline
		Methods &aero&bike&bird&boat&bottle&bus&car&cat&chair&cow&table&dog&horse&mbike&person&plant&sheep&sofa&train&tv&mAP\\
		\hline
		CNN-RNN &96.7&83.1&94.2&92.8&61.2  &82.1 & 89.1 & 94.2 & 64.2 & 83.6 & 70.0& 92.4& 91.7& 84.2& 93.7& 59.8& 93.2& 75.3& 99.7& 78.6& 84.0\\
		VGG+SVM & 98.9& 95.0& 96.8& 95.4& 69.7& 90.4 &93.5& 96.0 &74.2& 86.6& 87.8& 96.0& 96.3& 93.1& 97.2& 70.0& 92.1& 80.3 &98.1& 87.0 &89.7\\
		Fev+Lv& 97.9 &97.0& 96.6& 94.6& 73.6& 93.9& 96.5& 95.5& 73.7& 90.3& 82.8 &95.4 &97.7 &95.9& 98.6& 77.6& 88.7& 78.0 &98.3& 89.0& 90.6\\
		HCP & 98.6 & 97.1 & 98.0 & 95.6& 75.3& 94.7& 95.8& 97.3& 73.1& 90.2& 80.0& 97.3& 96.1& 94.9& 96.3& 78.3& 94.7& 76.2& 97.9& 91.5& 90.9\\	
		RDAL& 98.6& 97.4& 96.3& 96.2& 75.2& 92.4& 96.5& 97.1& 76.5& 92.0& 87.7& 96.8& 97.5& 93.8& 98.5& 81.6& 93.7& 82.8& 98.6& 89.3& 91.9\\
		RARL& 98.6& 97.1& 97.1& 95.5& 75.6& 92.8& 96.8& 97.3& 78.3& 92.2& 87.6& 96.9& 96.5& 93.6& 98.5& 81.6& 93.1& 83.2& 98.5& 89.3& 92.0\\
		SSGRL & 99.7& 98.4& 98.0& 97.6& 85.7& 96.2& 98.2& 98.8& 82.0& 98.1& 89.7& 98.8& 98.7& 97.0& 99.0& 86.9& 98.1& 85.8& 99.0& 93.7& 95.0\\
		MCAR& 99.7& 99.0& 98.5& 98.2& 85.4& 96.9& 97.4& 98.9& 83.7& 95.5& 88.8& 99.1& 98.2& 95.1& 99.1& 84.8& 97.1& 87.8& 98.3& 94.8& 94.8\\
		ASL(TResL)& 99.9 & 98.4 & 98.9& 98.7& 86.8& 98.2& 98.7& 98.5& 83.1& 98.3& 89.5& 98.8& 99.2& 98.6& 99.3& 89.5& 99.4& 86.8& 99.6& 95.2& 95.8\\
		ADD-GCN            & 99.8          & 99.0          & 98.4          & 99.0& 86.7& 98.1& 98.5& 98.3& 85.8& 98.3& 88.9& 98.8& 99.0& 97.4& 99.2& 88.3& 98.7& \textbf{90.7}& 99.5& \textbf{97.0} & 96.0\\
		Q2L-TResL          & 99.9          & 98.9          & 99.0          & 98.4& 87.7& 98.6& 98.8& 99.1& 84.5& 98.3& 89.2& 99.2& 99.2& 99.2& 99.3& 90.2& 98.8& 88.3& 99.5& 95.5& 96.1\\
		\hline
		PAT-T + TResL      & \textbf{100}  & 98.8          & 99.3          & 98.6& 88.8 & \textbf{99.0} & 98.8 & 99.1&  84.5&  98.9& 90.1 & 99.5& \textbf{99.4} &  98.6&  99.3&  89.2& 99.4 &  86.3& 99.7 &  95.8& 96.2\\
		\hline
		TResL* (21k)       & \textbf{100}  & 99.1          & \textbf{99.6} & 98.6 & 89.5 & 97.9 &  99.1 & \textbf{99.4} & 85.7 & \textbf{100} &  90.3 &  99.5 &
		99.2 & 99.5 & 99.3 & 90.9 & \textbf{99.9} & 88.9 & 99.8 & 96.3 & 96.6\\
		Q2L-TResL* (21k)   & \textbf{100}  & 99.2          &  99.5          & 98.8 & 88.7  & 98.9 & 98.9 & \textbf{99.4} & \textbf{87.3} & \textbf{100} &  89.8 & 99.3 &
		98.0 &  \textbf{99.6} & \textbf{99.5} & 91.1 & \textbf{99.9} & 90.6 & \textbf{99.9} & 96.3 &96.7 \\
		\hline
		PAT-T + TResL (21k)& \textbf{100}  & \textbf{99.5} & 99.5 & \textbf{99.4} & \textbf{90.1} & 98.4 & \textbf{99.2} & 99.3 & 86.8 & 99.9 & \textbf{90.5} & \textbf{99.7} & 99.3 & 99.3 & 99.4 & \textbf{91.6} & \textbf{99.9} & 89.8 & 99.8 & 96.8 & \textbf{96.9}\\
		\hline
	\end{tabular}
	\label{tb:voc_appendix}
\end{table*}

\begin{table*}[!t]
	\centering
	\caption{Comparison results on MS-COCO between our method and state-of-the-art methods. The backbones denoted with 21k and OI are pretrained on the ImageNet-21k and Open Image datasets, respectively. All metrics are in \%.}
	\begin{tabular}{c | c | c | c | c | c | c | c | c}
		\hline\hline
		Method  & Backbone & mAP   & CP     & CR    & CF1   & OP & OR & OF1 \\
		\hline
		ASL    & TResNetL       &  86.6 & 80.0 & 80.6 & 80.3 &	81.9 & 82.8 & 82.3 \\
		PAT-I      & TResNetL       &  87.6 & 87.9 &	76.5 &	81.8 &	89.3 &	79.1 &	83.9  
		\\
		PAT-T      & TResNetL       &  88.7 & 89.0 &	78.8 &	83.6 &	89.3 &	81.3 &	85.1 
		\\
		\hline
		TResNetL   & TResNetL (21k) &  88.4 & -    &  -   & -    & -    &  -   & - \\
		PAT-I  & TResNetL (21k) &  89.5 & 87.3 &	80.6 &	83.8 &	88.5 &	82.6 &	85.5 
		\\
		PAT-T  & TResNetL (21k) &  90.4 & 90.6 &	81.0 &	85.5 &	90.9 &	82.9 &	86.7 
		\\
		\hline
		Q2L-TResL  & TResNetL (21k)  &  89.2 & 86.3 & 81.4 & 83.8 & 86.5 & 83.3 & 84.9 \\
		PAT-I + Q2L-TResL  & TResNetL (21k)  &  89.8 & 85.7 &	81.1 &	83.3 &	87.4 &	82.4 &	84.8 
		\\
		PAT-T + Q2L-TResL  & TResNetL (21k) &  91.0 & 89.1 &	82.7 &	85.8 &	89.8 &	83.9 &	86.8 
		\\
		\midrule
		ML-Decoder  & TResNetL (OI)   &  90.0 & -    & -    & -     & -     &   -   & -    \\ 
		PAT-I + ML-Decoder& TResNetL (OI)   &  90.5 & 86.0 &	82.9 &	84.4 &	87.5 &	84.4 &	86.0  
		\\
		PAT-T + ML-Decoder& TResNetL (OI)   &  91.4 & 90.1 & 82.9 & 86.4  & 90.5 & 84.8 & 87.6  \\
		\hline\hline
	\end{tabular}
	\label{tb:coco_app}
\end{table*}

\section{Derivation of Eq.~\eqref{eq:tde2}}

\label{app:tde}

Given an image containing the target object and its co-occurring object $\{X=\x, Z=\z\}$, $\{X=\x, Z=\z_0\}$ and $\{X=\x_0,Z=\z\}$ form a partition, which are mutually exclusive and collectively exhaustive. According to the equation of total probability, we have
\begin{equation}
\begin{aligned}
  P(Y_{\o}^k=1|X=\x_0, Z=\z)P(X=\x_0, Z=\z)& = P(Y_{\o}^k=1|X=\x, Z=\z)P(X=\x, Z=\z)\\
  &\qquad- P(Y_{\o}^k=1|X=\x, Z=\z_0)P(X=\x, Z=\z_0).
\end{aligned}
\end{equation}
By dividing both sides by $P(X=\x_0,Z=\z)$, we obtain
\begin{equation}\label{eq:res}
	\begin{aligned}
		P(Y_{\o}^k=1|X=\x_0, Z=\z)& = \frac{P(X=\x, Z=\z)}{P(X=\x_0, Z=\z)}P(Y_{\o}^k=1|X=\x, Z=\z)\\
		&\qquad- \frac{P(X=\x, Z=\z_0)}{P(X=\x_0, Z=\z)}P(Y_{\o}^k=1|X=\x, Z=\z_0)\\
		&= \alpha P(Y_{\o}^k=1|X=\x, Z=\z)-\beta P(Y_{\o}^k=1|X=\x, Z=\z_0),
	\end{aligned}
\end{equation}
where $\alpha=\frac{P(X=\x, Z=\z)}{P(X=\x_0, Z=\z)}$ and $\beta=\frac{P(X=\x, Z=\z_0)}{P(X=\x_0, Z=\z)}$ are data-dependent parameters.

Substituting the result of equation Eq.~\eqref{eq:res} into Eq.~\eqref{eq:tde}, we have
\begin{equation}
	\begin{aligned}
		\text{TDE}_k(\i)&=P(Y_{\o}^k=1|X=\x, Z=\z)-P(Y_{\o}^k=1|X=\x_0, Z=\z)\\
		&=(1-\alpha)P(Y_{\o}^k=1|X=\x, Z=\z)+\beta P(Y_{\o}^k=1|X=\x, Z=\z_0).
	\end{aligned}
\end{equation}
By dividing both sides by $1-\alpha$ for each example, we obtain $\text{TDE}_k(\i)=P(Y_{\o}^k=1|X=\x, Z=\z)+\lambda P(Y_{\o}^k=1|X=\x, Z=\z_0)$, where $\lambda=\beta/ (1-\alpha)$. We neglect the denominator $(1-\alpha)$ for $\text{TDE}_k(\i)$ since it does not affect the final result.

\begin{algorithm}[!t]
	\caption{Pseudocode of PAT-I in a PyTorch-like style}
	\label{alg:PAT-I}
	\begin{algorithmic}
			\STATE \texttt{\textcolor{teal}{\# f: model}}	
			\STATE \texttt{\textcolor{teal}{\# t: temperature}}	
			\STATE \texttt{\textcolor{teal}{\# lam: tde parameter}}	
			\STATE \texttt{for [x; u\_1; u\_2; u\_3; u\_4], y in loader:}
			\begin{ALC@g}
					\STATE {\textcolor{teal}{\texttt{\# image logits and patch logits NxCx5}}}	
					\STATE \texttt{[l\_x; l\_u\_1; l\_u\_2; l\_u\_3; l\_u\_4] = f([x; u\_1; u\_2; u\_3; u\_4])}
					\STATE {\textcolor{teal}{\texttt{\# patch weights NxCx4}}}
					\STATE \texttt{[w\_1; w\_2; w\_3; w\_4]=softmax([l\_u\_1;l\_u\_2;l\_u\_3;l\_u\_4]/t, dim=-1)}
					
					\STATE {\textcolor{teal}{\texttt{\# tde logits}}}
					\STATE \texttt{l\_tde = l\_x + lam*(w\_1*l\_u\_1 + w\_2*l\_u\_2 + w\_3*l\_u\_3 + w\_4*l\_u\_4).sum(dim=-1)}
					
					\STATE
					\STATE {\textcolor{teal}{\texttt{\# mAP score calculation}}}
					\STATE \texttt{mAP\_score = mAP\_function(l\_tde, y)}
				\end{ALC@g}
		\end{algorithmic}
\end{algorithm}

\begin{algorithm}[!t]
	\caption{Pseudocode of PAT-T in a PyTorch-like style}
	\label{alg:PAT-T}
	\begin{algorithmic}
		\STATE \texttt{\textcolor{teal}{\# f: model}}	
		\STATE \texttt{\textcolor{teal}{\# t: temperature}}	
		\STATE \texttt{for [x; u\_1; u\_2; u\_3; u\_4], y in loader:}
		\begin{ALC@g}
			\STATE {\textcolor{teal}{\texttt{\# image logits and patch logits NxCx5}}}	
			\STATE \texttt{[l\_x; l\_u\_1; l\_u\_2; l\_u\_3; l\_u\_4] = f([x; u\_1; u\_2; u\_3; u\_4])}
			\STATE {\textcolor{teal}{\texttt{\# patch weights NxCx4}}}
			\STATE \texttt{[w\_1; w\_2; w\_3; w\_4]=softmax([l\_u\_1;l\_u\_2;l\_u\_3;l\_u\_4]/t, dim=-1)}
			\STATE {\textcolor{teal}{\texttt{\# weighted logits}}}
			\STATE \texttt{l\_w = (w\_1*l\_u\_1 + w\_2*l\_u\_2 + w\_3*l\_u\_3 + w\_4*l\_u\_4).sum(dim=-1)}
			\STATE
			\STATE {\textcolor{teal}{\texttt{\# loss, Eq.(5)}}}
			\STATE \texttt{loss = AsymmetricLoss(l\_x, y) + AsymmetricLoss(l\_w, y)} 
			\STATE {\textcolor{teal}{\texttt{\# model update}}}
			\STATE \texttt{loss.backward()} 
		\end{ALC@g}
	\end{algorithmic}
\end{algorithm}

\section{Detailed Experimental Settings and Additional Results}

On VOC 2007 dataset, our method is compared with the following representative algorithms: CNN-RNN \cite{wang2016cnn}, VGG+SVM \cite{simonyan2014very}, Fev+Lv \cite{yang2016exploit}, HCP \cite{wei2015hcp}, RDAL \cite{wang2017multi}, RARL \cite{chen2018recurrent}, SSGRL \cite{chen2019learning}, and MCAR \cite{gao2021learning}.

\label{sec:moreres}
Table \ref{tb:voc_appendix} reports additional results based on TResNet (21k) on VOC 2007. We can see that our method outperforms the state-of-the-art methods. Table \ref{tb:coco_app} reports the additional results of PAT-I on MS-COCO. As a comparison, we also report the results of baseline methods and PAT-T. From the table, we can see that PAT-I can achieve better performance than baseline methods, which verifies that the issue of overfitting to co-occurrence relationships can be addressed by incorporating counterfactual inference via the patching-based strategy. It is noteworthy that PAT-I does not even require training the model, which means that we can flexibly apply it to any trained model.

\section{Pseudocode of PAT-I and PAT-T}
\label{sec:pseudocode}
Algorithm \ref{alg:PAT-I} and Algorithm \ref{alg:PAT-T} provide the pseudocode of PAT-I and PAT-T in a Pytorch-like style, respectively. We can see that both two algorithms are simple to implement and act like plug-and-play plugins, which bring clear benefits.

\end{document}